\documentclass[10pt,journal,compsoc]{IEEEtran}
\usepackage{times}
\usepackage[table]{xcolor}

\usepackage{helvet}
\usepackage{courier}
\usepackage{stfloats}
\usepackage{algorithmicx}
\usepackage{algorithm}
\usepackage{tikz}
\usetikzlibrary{decorations.pathreplacing}
\usepackage{amsmath}
\usepackage{arydshln}
\usepackage{enumitem}
\usepackage{algpseudocode}
\usepackage{multirow}
\usepackage{epsfig}
\usepackage{multicol,lipsum,xparse}
\usepackage{subcaption}
\usepackage{amsmath}
\usepackage{stmaryrd}
\usepackage{color}
\usepackage{todonotes}
\usepackage[utf8]{inputenc}
\usepackage{tikz}
\ifCLASSOPTIONcompsoc
  \usepackage[nocompress]{cite}
\else
  \usepackage{cite}
\fi

\begin{document}

\title{Performance Estimation of Synthesis Flows cross Technologies using LSTMs and Transfer Learning}

\author{Cunxi~Yu~\IEEEmembership{ Member,~IEEE,}
        and Wang~Zhou~
\IEEEcompsocitemizethanks{\IEEEcompsocthanksitem C. Yu is with the Department
of Electrical and Computer Engineering, Cornell University, Ithaca, New York, 14850, USA. W. Zhou is with IBM Thomas J. Watson Research Center, Yorktown Height, New York, 10598, USA.\protect\\
E-mail: cunxi.yu@cornell.edu, wang.zhou@ibm.com}}
\maketitle

\newcommand{\pgftextcircled}[1]{
    \setbox0=\hbox{#1}%
    \dimen0\wd0%
    \divide\dimen0 by 2%
    \begin{tikzpicture}[baseline=(a.base)]%
        \useasboundingbox (-\the\dimen0,0pt) rectangle (\the\dimen0,1pt);
        \node[circle,draw,outer sep=0pt,inner sep=0.1ex] (a) {#1};
    \end{tikzpicture}
}

\begin{abstract}
Due to the increasing complexity of Integrated Circuits (ICs) and System-on-Chip (SoC), developing high-quality synthesis flows within a short market time becomes more challenging. We propose a general approach that precisely estimates the Quality-of-Result (QoR), such as delay and area, of unseen synthesis flows for specific designs. The main idea is training a Recurrent Neural Network (RNN) regressor, where the flows are inputs and QoRs are ground truth. The RNN regressor is constructed with Long Short-Term Memory (LSTM) and fully-connected layers. This approach is demonstrated with 1.2 million data points collected using 14nm, 7nm regular-voltage (RVT), and 7nm low-voltage (LVT) FinFET technologies with twelve IC designs. The accuracy of predicting the QoRs (delay and area) within one technology is $\boldsymbol{\geq}$\textbf{98.0}\% over $\sim$240,000 test points. To enable accurate predictions cross different technologies and different IC designs, we propose a transfer-learning approach that utilizes the model pre-trained with 14nm datasets. Our transfer learning approach obtains estimation accuracy $\geq$96.3\% over $\sim$960,000 test points, using only 100 data points for training.
\end{abstract}


\section{Introduction}

Electronic Design Automation (EDA) is important due to the increasing complexity of the designs and technologies. Demands for designing electronic systems in novel application domains, such as neuromorphic chips \cite{akopyan2015truenorth} and deep learning chips \cite{jouppi2017datacenter}, raise the challenges to a new level. For example, due to the lack of predictability of EDA techniques, the expensive design iterations with extensive human supervision are unavoidable. Developing and tuning design flows for specific designs is very time-consuming. In addition, most design flows are currently developed based on the knowledge of designers with iterative testing. However, because of the huge search space of flows, it is difficult to know how good the developed flows are. Hence, predictive flow-level modeling and prediction, and design-specific tuning have very high value \cite{kahng2018new}\cite{burns2018keynote}.

\textit{Deep learning} has shown considerable success in a broad area of applications using various deep neural network architectures, such as Convolutional Neural Network (CNN) and Recurrent Neural Network (RNN). A CNN, which is one type of feed-forward artificial neural networks constructed with convolutional and pooling layers, has shown great performance in analyzing images \cite{krizhevsky2012imagenet}\cite{girshick2015fast}. An RNN is an artificial neural network where the connections between units form a directed graph along a sequence. This allows it to exhibit dynamic temporal behavior for a timed sequence. Unlike feed-forward neural networks, RNNs can use their internal memories to process sequences of inputs. This makes them applicable to tasks such as speech recognition \cite{graves2013speech}, language translation \cite{bahdanau2014neural}, and generating textual descriptions \cite{vinyals2015show}. Deep learning has also been used for EDA techniques. For example, ResNet \cite{kahng2018new} has been used for lithography modeling optimization \cite{lin2018data}, and CNN has been used as a flow classifier for generating design-specific synthesis flows \cite{cunxi2018CNN}.  

In this paper, an RNN regression model is proposed to estimate the performance of flows. The approach is demonstrated by predicting the Quality-of-Results of logic synthesis flows with three different technologies. Furthermore, this approach can be used for performance estimation of flows in different domains, such as physical design flows, compiling flows, etc. The main contributions of this paper are:

\begin{itemize}
\item A closed formula is introduced to represent the search space for arbitrary types of flows. A synthesis embedding model that represents flows as discrete sequences using 2-D matrix is introduced.
\item An LSTM based RNN regression architecture is proposed. The inputs are flows in the timed-model matrix, and ground truth are delay and area collected after technology mapping.
\item We propose a transfer-learning approach that adapts the model learned from one technology node to another technology node. This offers the ability to estimate the performance for next/future technology nodes.
\item The approach has been demonstrated with $\sim$1.2 million data points with 14nm, regular-voltage (RVT) 7nm, and low-voltage (LVT) 7nm FinFET technologies, collected with 12 different IC designs. We achieve testing accuracy $\geq$98.0\% for specific design and technology, and testing accuracy cross technologies and designs is $\geq$96.3\% after transfer learning.
\item We demonstrate that the LSTM based approach with transfer learning outperforms the CNN based approach \cite{cunxi2018CNN} with all the datasets using only 25 training points.
\end{itemize}

\section{Background}

\subsection{Synthesis flows and Search Space}\label{sec:synthesis_flow}

Synthesis flows are a set of synthesis transformations that apply iteratively to the input designs. The synthesis transformations are mainly involved in three stages of the design flow: high-level synthesis (HLS), logic synthesis (LS) and placement and route (PnR). For different types of electronic designs, the flows need to be changed accordingly. 

In general, there are two types of flows, \textit{none-repetition} flows and \textit{m-repetition} flows \cite{cunxi2018CNN}. Given $n$ unique transformations, a flow developed with these transformations is called none-repetition flow if each transformation appears only once. The length of none-repetition flows is $n$. For $m$-repetition flows, each transformation appears $m$ times. The length of $m$-repetition flows is $m$$\cdot$$n$. In \cite{cunxi2018CNN}, the upper bound of the search space for both types of flows are discussed. For none-repetition flows, a closed representation $n!$ is the upper bound of its search space. An iterative formula is used to describe the search space of $m$-repetition. However, the upper bound was given as a range without a closed formula representation. 

A closed formula is introduced to describe the search space of repetition flows. The search space for $m$-repetition flows is a \textit{multiset} permutation problem. Specifically, for $m$-repetition flows with $n$ unique transformations, the search space is shown in Equation 1. 
\begin{align}
\small
\dfrac{(n\cdot m)!}{(m!)^n}
\end{align}

Using the multiset permutation concept, we generalize the formula to describe the search space for any type of flows. Let $n$ be the number of unique transformation, the $M$-repetition flows, $M$=\{$m_{1}, m_{2}, ...,m_{n}$\}, where $m_i$ is the number of repetitions of the $i^{th}$ transformation. The total number of possible flows is shown in Equation 2. 
\begin{align}
\small
S(m,n) = \dfrac{(m_1+m_2+\cdot\cdot\cdot m_n)!}{(m_1!)(m_2!)\cdot\cdot\cdot(m_n!)}
\end{align}

\subsection{Quality of Result of Digital ICs}

\begin{figure}[!htb]
\centering
\begin{minipage}{0.24\textwidth}
  \centering
\includegraphics[width=1\textwidth]{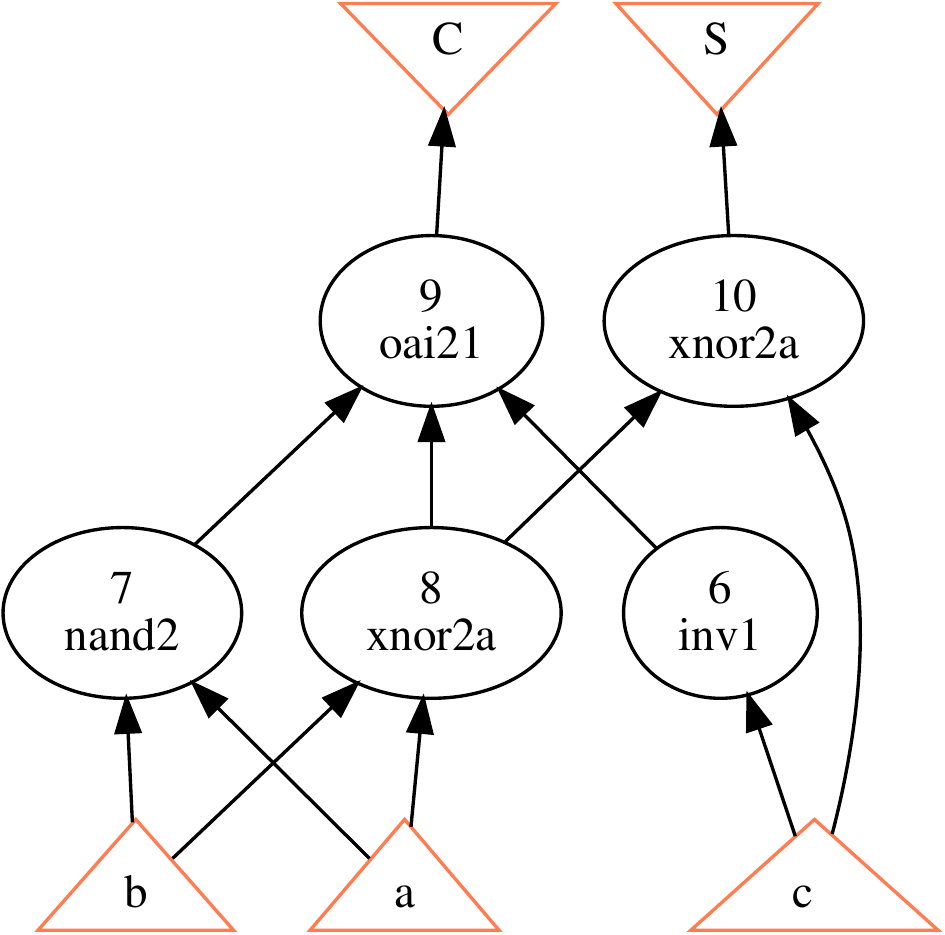}
\subcaption{}\label{fig:fa_labeling_a}
\end{minipage}%
\begin{minipage}{0.2\textwidth}
  \centering
\includegraphics[width=1\textwidth]{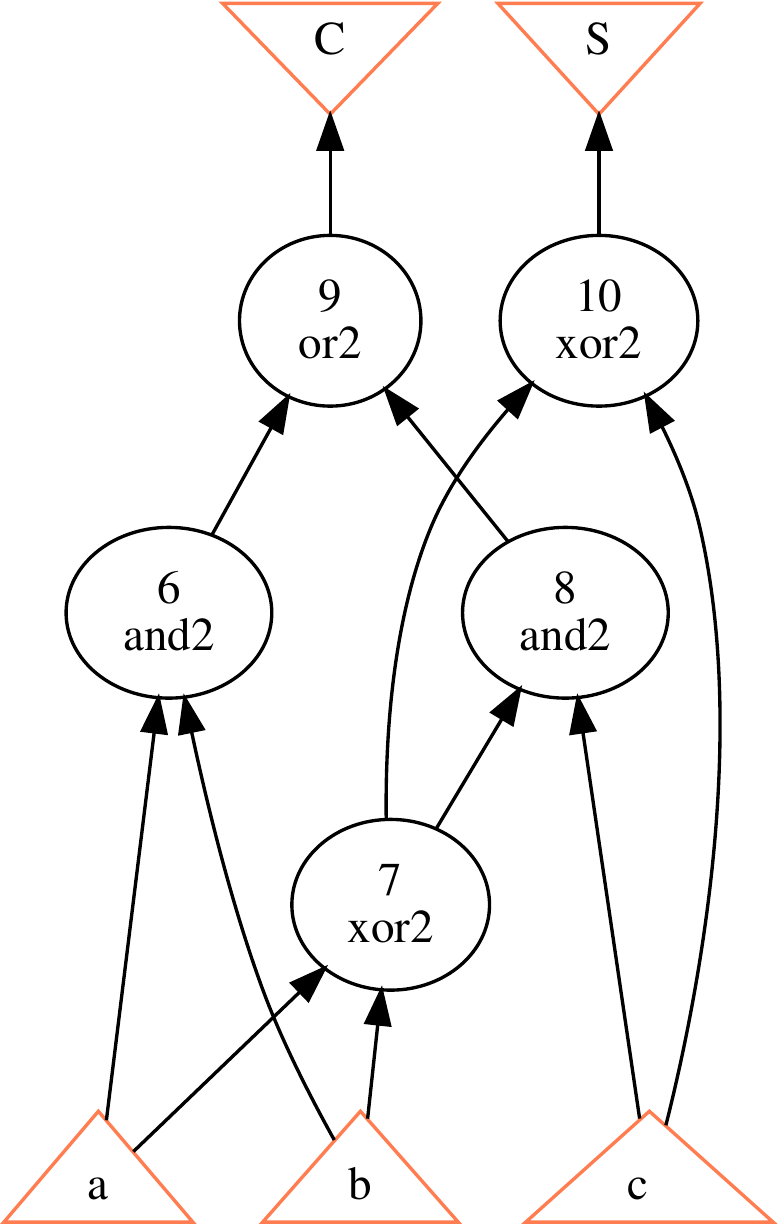}
\subcaption{}\label{fig:fa_labeling_b}
\end{minipage}%
\caption{1-bit Full Adder ($F= 2C+S=a+b+c$) gate-level netlists produced by two different synthesis flows using 7nm FinFET technology library. a) Delay = 30.1 ps, Area = 9.40 um$^2$; b) Delay = 47.6 ps, Area = 8.38 um$^2$.}
\label{fig:full_adder}
\end{figure}

Quality of Results (QoR) of Integrated Circuits (IC) is a term used in evaluating technological processes. Mostly, it is represented as a vector of values that describes the performance of the designs and design process. For example, a QoR could include the critical path delay (chip frequency = 1$/$Delay), power, area, e.g., \{1 ns (1 GHz), 2.5 W, 1 mm$^2$\}. Given a design specification, the QoR of the technological processed designs can be very different using different synthesis flows. An illustrative example of mapping a 1-bit Full Adder using 7nm FinFET technology library is shown in Figure \ref{fig:full_adder}. The specification of a 1-bit Full Adder is $a+b+c = 2C + S$, where $a,b,c,C,S$ are binary signals. Using two different synthesis flows, two gate-level netlists are produced. In Figure \ref{fig:full_adder}, each node represents one logic gate and the type label of the node is the type of the logic gate. Note that the performance of different logic gates are different and they are defined based on the technology information. For example, delay and area of XOR2 are 15.95 ps and 2.8 um$^2$; delay and area of NAND2 are 11.90 ps and 1.39 um$^2$. After applying synthesis flow, we observe that the QoRs of two netlists are different. The design in Figure \ref{fig:fa_labeling_a} is about 37\% faster than design in Figure \ref{fig:fa_labeling_b}, and requires 12\% more area. Note that the QoR can only be obtained after the entire technological design process. For large designs, this process is extremely time-consuming.

Depending on the application, the design objectives can be very different. For instance, for high-performance image processing designs, the designs are mostly designed to be as fast as possible, i.e., delay to be as small as possible. For Internet-of-Things (IoT) designs, the power and area are required to be minimized. The massive search space of synthesis flows (Equations 1 and 2) and the time-consuming technological design process, along with the various design objectives, are the main motivations of this work.

\subsection{Recurrent neural network}\label{sec:rnn_lstm}

Recurrent Neural Network (RNN) is a class of artificial neural network with a chain of units in a directed graph sequence. RNNs perform the same computations for each unit in a sequence, and output states depend on the previous states. In theory, RNNs can make use of information in arbitrarily long sequences, but in practice, they are limited to looking back only a few steps, called ``Long-Term Dependencies'' problem \cite{hochreiter1991untersuchungen}. Long Short-Term Memories (LSTMs) \cite{hochreiter1997long} are explicitly designed to address the long-term dependency problem by adding control gates to the recurrent units. Such controlled states are referred to as a gated state or gated memory, which have been implemented as part of LSTMs and Gated Recurrent Units (GRUs)\cite{cho2014learning}. An RNN composed of LSTM units is often called an LSTM network. A common LSTM unit is composed of a cell, an input gate, an output gate and a forget gate. The cell is responsible for ``remembering'' values over arbitrary time intervals. 


The LSTM has the ability to remove or add information to the cell state, carefully regulated by structures called gates. Gates are a way to optionally let information through. They are composed out of a sigmoid neural net layer and a pointwise multiplication operation. The sigmoid gate (layer) outputs numbers in the range of $[0,1]$, which determines the ratio of the outputs of previous component going into next one. 



\subsection{Related work}

Yu et al. \cite{cunxi2018CNN} presented a deep learning based approach for generating design-specific synthesis flows for ABC. The main idea of this approach is formulating the flow optimization problem as a \textit{Multiclass Classification} problem. The authors proposed to use Convolutional Neural Network (CNN) based classifier that includes two \textit{Convolution+MaxPool} layers and three \textit{Dense} layers. It shows that the classifier can successfully distinguish the best and worst flows given the design objectives. However, there are two main limitations:
\begin{itemize}
    \item The classifier can only classify the flows into different performance classes, however, it cannot distinguish the performance of different flows within the same class.
    \item The prediction accuracy heavily relies on the labeling rules since the labels of the flows are post-created based on the Quality-of-Result (QoR).
\end{itemize}

\begin{figure}[!htb]
\centering
\begin{minipage}{0.24\textwidth}
  \centering
\includegraphics[width=1\textwidth]{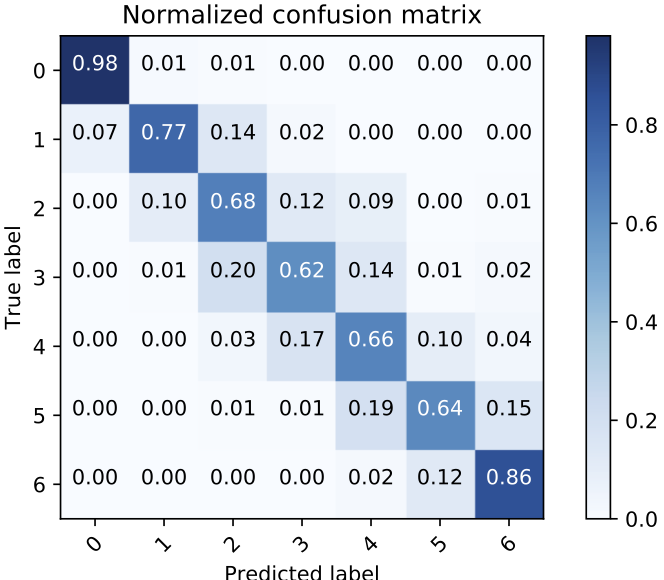}
\subcaption{Labeling rule 1.}\label{fig:labeling_a}
\end{minipage}%
\begin{minipage}{0.24\textwidth}
  \centering
\includegraphics[width=1\textwidth]{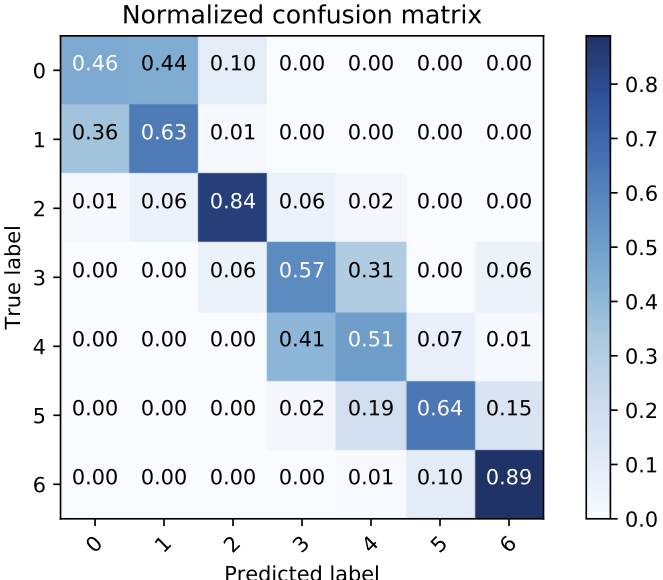}
\subcaption{Labeling rule 2.}\label{fig:labeling_b}
\end{minipage}%
\caption{Confusion matrices of two classifiers that trained with two different labeling rules.}
\label{fig:labeling_rule}
\end{figure}

It is obvious that the first limitation comes from the idea of flow classification. Therefore, we focus on illustrating the second limitation. The single-metric and multi-metric rules are introduced in \cite{cunxi2018CNN} that label the synthesis flows based on single QoR or multiple QoR metrics, such as area, delay, etc. The labeling rule for \textit{seven}-classes labeling requires \textit{six} QoR delimiters. For example, let the six delimiters be the data point at 7\%, 20\%, 40\%, 65\%, 80\%, and 93\% position of training datasets (assuming the training set is sorted from best-to-worse QoR), namely \textit{Labeling rule 1} (Figure \ref{fig:labeling_a}). Alternatively, let the six delimiters be the data point at 5\%, 15\%, 40\%, 65\%, 90\%, and 95\% position of training datasets, namely \textit{Labeling rule 2} (Figure \ref{fig:labeling_b}). We compare the classification performance of two labeling rules using the CNN architecture and the 64-bit Montgomery Multiplier dataset proposed in \cite{cunxi2018CNN}. The training and testing of CNN Classifiers are done with Keras using Tensorflow as backend. The results are shown in the confusion matrices in Figure \ref{fig:labeling_rule}.

As mentioned in \cite{cunxi2018CNN}, the main tasks are searching for the best (class 0) and worst (class 6) flows for a given design objective. The results show that labeling rule 1 provides 98\% accuracy for predicting class 0, and 86\% accuracy for class 6; labeling rule 2 provides only 46\% accuracy for class 0, and 89\% for class 6. We can see that using different labeling rules, the performance of the classifiers can be very different. Note that the labeling rule input to this approach, which should be defined by the user. These two limitations offer the main motivation for our regression based approach. 

As described previously, a synthesis flow is a \textbf{sequence} of transformations. Mostly in IC design, QoR of a large number of synthesized designs is a continuous variable. To achieve accurate estimation of a continuous variable that is related to sequential behaviors, we explore LSTM network regressor in this work. In the result section, we also compare the performance with a regressor built using the proposed CNN model \cite{cunxi2018CNN}.

\section{Modeling}

This section introduces the inputs of the neural network and ground truth for the regression model. The inputs are the synthesis flows that are represented by a 2-D matrix using a novel \textit{Timed-Model} of flows. The ground truth includes delay and area results after technology mapping.

\subsection{Inputs: Timed-Model of flows}

\vspace{0mm}
\begin{table}[!htb]
\caption{Illustration of timed-model of synthesis flows using ABC default synthesis flow \textit{resyn}. Synthesis transformation at each time spot is shown above the visualized binary matrix.}
\small
\centering
\begin{tikzpicture}[scale=0.65]
\node[align=center] at (1,0.85) {balance};
\node[align=center] at (2,1.5) {$t_1$};
\draw [thick,->] (2,1.2) -- (2,0.35);
\draw [thick,decorate,decoration={brace,amplitude=6pt,raise=0pt}] (0,0.15) -- (2,0.15);
\node[align=center] at (3.15,-0.85) {rw};
\node[align=center] at (4.15,-1.5) {$t_2$};
\draw [thick,->] (4.15,-1.2) -- (4.15,-0.35);
\draw [thick,decorate,decoration={brace,amplitude=6pt,raise=0pt,mirror}] (2.15,-0.15) -- (4.15,-0.15);
\node[align=center] at (5.5,0.85) {rwz};
\node[align=center] at (6.3,1.5) {$t_3$};
\draw [thick,->] (6.3,1.2) -- (6.3,0.35);
\draw [thick,decorate,decoration={brace,amplitude=6pt,raise=0pt}] (4.3,0.15) -- (6.3,0.15);
\node[align=center] at (7.45,-0.85) {balance};
\node[align=center] at (8.45,-1.5) {$t_4$};
\draw [thick,->] (8.45,-1.2) -- (8.45,-0.35);
\draw [thick,decorate,decoration={brace,amplitude=6pt,raise=0pt,mirror}] (6.45,-0.15) -- (8.45,-0.15);
\node[align=center] at (9.6,0.85) {rwz};
\node[align=center] at (10.6,1.5) {$t_5$};
\draw [thick,->] (10.6,1.2) -- (10.6,0.35);
\draw [thick,decorate,decoration={brace,amplitude=6pt,raise=0pt}] (8.6,0.15) -- (10.6,0.15);
\node[align=center] at (11.75,-0.85) {balance};
\node[align=center] at (12.75,-1.5) {$t_6$};
\draw [thick,->] (12.75,-1.2) -- (12.75,-0.35);
\draw [thick,decorate,decoration={brace,amplitude=6pt,raise=0pt,mirror}] (10.75,-0.15) -- (12.75,-0.15);
\draw [thick,->] (0,0) -- (12.8,0);
\end{tikzpicture}
\begin{tabular}{|c|c|c|c|c|c|}
\hline
$t_1$ & $t_2$ & $t_3$ & $t_4$ & $t_5$ & $t_6$ \\ \hline\hline
$b$ & $rw$ & $rwz$ & $b$ & $rwz$ & $b$ \\ \hline\hline
\cellcolor[HTML]{333333} &  &  & \cellcolor[HTML]{333333} &  & \cellcolor[HTML]{333333} \\ \hline
 & \cellcolor[HTML]{333333} &  &  &  &  \\ \hline
 &  & \cellcolor[HTML]{333333} &  & \cellcolor[HTML]{333333} &  \\ \hline
\end{tabular}
\label{fig:timed_model_fows}
\end{table}

\begin{table*}[t]
\centering
\scriptsize
\caption{Example of a 24 transformations long synthesis flow using the presented timed-model. The flow is a $4$-repetition synthesis flow with six unique transformations, \{{\it{rw, rwz, b, rs, rfz, rf}}\}, and each transformation repeats four times. The complete synthesis flow is shown in the second row and synthesis transformation at each time spot is shown above the visualized binary matrix.}
\begin{tabular}{|l|l|l|l|l|l|l|l|l|l|l|l|l|l|l|l|l|l|l|l|l|l|l|l|l|l|l|l|}
\hline
$t_1$ & $t_2$ & $t_3$ & $t_4$ & $t_5$ & $t_6$ & $t_7$ & $t_8$ & $t_9$ & $t_{10}$ & $t_{11}$ & $t_{12}$ & $t_{13}$  & $t_{14}$ & $t_{15}$ & $t_{16}$ & $t_{17}$ & $t_{18}$ & $t_{19}$ & $t_{20}$ & $t_{21}$ & $t_{22}$ & $t_{23}$ & $t_{24}$ \\ \hline\hline
b & rf & rwz & rw & rs & rfz & b & rw & rwz & rf & rs & rfz & b  & rw & rwz & rw & rs & rfz & b & rf & rwz & rfz & rs & rf \\ \hline\hline
  &  &   & \cellcolor[HTML]{333333}  &   &   &   & \cellcolor[HTML]{333333} &   &   &   &   &   & \cellcolor[HTML]{333333} &   & \cellcolor[HTML]{333333}  &   &   &   &  &   &   &   &      \\ \hline
  &   & \cellcolor[HTML]{333333} &   &   &   &   &   & \cellcolor[HTML]{333333} &   &   &   &   &   & \cellcolor[HTML]{333333} &   &   &   &   &   & \cellcolor[HTML]{333333} &   &   &    \\ \hline
  \cellcolor[HTML]{333333} &   &   &   &   &   & \cellcolor[HTML]{333333} &   &   &   &   &   & \cellcolor[HTML]{333333} &   &   &   &   &   & \cellcolor[HTML]{333333} &   &   &   &   &    \\ \hline
  &   &   &   & \cellcolor[HTML]{333333} &   &   &   &   &   & \cellcolor[HTML]{333333} &   &   &   &   &   & \cellcolor[HTML]{333333} &   &   &   &   &   & \cellcolor[HTML]{333333} &    \\ \hline
  &   &   &   &   & \cellcolor[HTML]{333333} &   &   &   &   &   & \cellcolor[HTML]{333333} &   &   &   &   &   & \cellcolor[HTML]{333333} &   &   &   & \cellcolor[HTML]{333333}  &   &   \\ \hline
    & \cellcolor[HTML]{333333}  &   &  &   &   &   &   &   & \cellcolor[HTML]{333333} &   &   &   &   &   &  &   &   &   & \cellcolor[HTML]{333333}  &   &  &   & \cellcolor[HTML]{333333}   \\ \hline
\end{tabular}
\label{fig:model_flows_24}
\end{table*}

As mentioned earlier, any flow includes a set of transformations that perform iteratively. We illustrate the concept of timed-model using one common logic synthesis flow provided in ABC \cite{mishchenko2010abc}, \textit{\textbf{resyn}}, which includes six transformations: \textit{balance (b), rewrite (rw), rewrite -z (rwz), b, rwz, b}. 

The time-line of applying \textit{\textbf{resyn}} to designs is shown in Table \ref{fig:timed_model_fows}. Each transformation in this flow is applied to the design at each time frame. For example, for time in range (0,$t_1$), the transformation \textit{balance} is applied and it finishes at $t_1$. Then, the second transformation \textit{rewrite} starts and finishes at $t_2$. The whole flow finishes at $t_6$. Note that the runtime of different transformations can be very different; and the runtime of the same transformation at different stage could be different as well. In this work, the runtime of each transformation is not included in the modeling. This means that the timed-model of the flows is considered as a discrete sequence. Using one-hot encoding for the three transformations in \textit{resyn}, let \textit{balance}=[1 0 0], \textit{rw} = [0 1 0], and \textit{rwz} = [0 0 1]. The resulting timed-model of \textit{resyn} in binary matrix is shown in Table \ref{fig:timed_model_fows}.

A more complex example is shown in Table \ref{fig:model_flows_24}. The input synthesis flow is an ABC synthesis flow including six transformations, \textit{balance (b), restructure (rs), rewrite (rw), refactor (rf), rewrite -z (rwz), refactor -z (rfz)}, and these transformations are repeated four times. The length of this synthesis flow is 24 such that it requires 24 time-frames to complete. With total six transformations, the final model of this synthesis flow is a matrix of shape (6,24) (Figure \ref{fig:model_flows_24}). This type of matrices will be the input to the neural network for training and inference.

\subsection{Ground truth}

In the context of machine learning, the ground truth is a measurement of the target variable(s) for the training and testing data points. In other words, the ground truth defines the objective(s) of the learning model. In the scenario of training a regressor for synthesis, taking synthesis flows as inputs, the ground truth could be \textit{synthesis runtime}, \textit{critical path delay}, \textit{total logic area}, \textit{XOR counts}, etc. Similarly, this can be extended to other flow performance estimation problem such as placement and route, with the ground truth being \textit{worst negative slacks}, \textit{total negative slacks}, \textit{routing length}, etc. In the result of this paper, the demonstration and evaluation of the proposed approach specifically target on synthesis flows of the open source logic synthesis framework ABC \cite{mishchenko2010abc}. The ground truth includes \textit{critical path delay} (delay in short) and \textit{logic area} (area in short).

\section{Approach}

This section presents the implementation of LSTM based RNN regressor, training setup and the summary of datasets.

\subsection{LSTM network architecture}

The RNN regressor architecture is presented in Table \ref{tbl:table_rnn_model}. The regressor is designed with LSTM$\times$2, Batch Normalization (BN)$\times$4, Dropout$\times$1, and Dense layers$\times$3. The first column shows the layers and its type in a top-down order. The second column presents the output shape of the current layer, and the last column shows the number of parameters in each layer. The activation function of the Dense1 and Dense2 layers is \textit{ReLu}. The output layer is implemented with a dense layer where the number of units equals to the ground truth dimension, $dim$. In this work, the ground truth dimension is one, i.e., either area or delay. The activation function for the last layer is \textit{Linear}.


\begin{itemize}
\item \textbf{LSTM Layer (Layers 1 and 3):} The core of the model consists of two LSTM layers. Both LSTM layers include 128 hidden units. The inner recurrent activation applied to input, forget, and output gates is \textit{hard sigmoid}, i.e., segment-wise linear approximated \textit{sigmoid} function. The activation for the hidden state and output hidden state is hyperbolic tangent function ({tanh}).
\end{itemize}

\begin{itemize}
\item \textbf{Batch Normalization (Layers 2,4,6,8):} Batch Normalization (BN) \cite{ioffe2015batch} in general helps the training in speed and accuracy. The basic idea of batch normalization is similar to data normalization in training data pre-processing. Instead of applying normalization to the training data only, BN applies normalization over the hidden layers. 
\end{itemize}

\begin{itemize}
\item \textbf{Dropout Layers (Layer 9):} Dropout is a regularization technique, which aims to reduce the complexity of the model with the goal to prevent overfitting. Dropout layer consists in randomly setting a fraction rate of input units to 0 at each update during training time \cite{srivastava2014dropout}. The units that are kept are scaled by $1/(1 - r)$, where $r$ is the dropout rate, so that the sum is unchanged during training and inference processes. In this paper, the dropout rate is 0.4.
\end{itemize}

\begin{itemize}
\item \textbf{Dense Layers (Layers 5,7,10):} Dense layer is applying linear operation in which every input is connected to every output by a weigh, generally followed by a non-linear activation function to add nonlinearity to the model. Specifically, the dense layer in this work performs \textit{activation(multiply(input, kernel))}, where \textit{activation} is the element-wise activation function, kernel is a weights matrix created by the layer. The values in the kernel matrix are the trainable parameters which get updated during back-propagation. The activation functions of the Dense1 and Dense2 layers is \textit{ReLu}, and activation of last layer is \textit{Linear}. 
\end{itemize}

\begin{table}[!htb]
\centering
\small
\caption{LSTM based RNN model architecture, including the output shape and number of parameters of each layer.}
\begin{tabular}{|l|l|r|}
\hline
\textbf{Layer : Type}        & \textbf{Output Shape} & \textbf{\# Param} \\ \hline\hline
1:LSTM1          & (None, 24, 128)  & 68608    \\ \hline
2:BN1 & (None, 24, 128)  & 512      \\ \hline
3:LSTM2          & (None, 128)  & 131584    \\ \hline
4:BN2 & (None, 128)  & 512      \\ \hline
5:Dense1              & (None, 30)   & 3870     \\ \hline
6:BN3 & (None, 30)   & 120       \\ \hline
7:Dense2              & (None, 30)   & 930     \\ \hline
8:BN4 & (None, 30)   & 120       \\ \hline
9:Dropout              & (None, 30)   & 0        \\ \hline
10:Dense3             & (None, $dim$)    & 31$\times dim$      \\ \hline
\end{tabular}
\label{tbl:table_rnn_model}
\end{table}

\subsection{Datasets}

The datasets are generated by logic synthesis tool ABC \cite{mishchenko2010abc}, with 100,000 random flows generated. All 100,000 flows are applied to three different designs, 64-bit Montgomery Multiplier, 64-bit ALU and 128-bit AES core, using 14nm, 7nm RVT and 7nm LVT technologies. For exploring the transferability cross designs and technologies, we apply the first 20,000 random flows to nine more designs with different IPs (intellectual property) (Table \ref{tbl:datasets}), including cryptographic hash {\it SHA}, RISC (Reduced Instruction Set Computer) architecture Open RISC 1200 \textit{OR1200}, etc. These designs are obtained from OpenCore \cite{opencore-web}. Note that some of the random flows fail because of the internal ABC crashes (segment fault reported). There are three failure cases observed while applying to the Montgomery multiplier, and 263 failure cases for the AES core. There are $\sim$300,000 data points generated using 14nm technology with 3 designs, and $\sim$960,000 data points using 7nm technologies with 12 designs. The summary of the datasets is shown in Table \ref{tbl:datasets}.

\noindent
\textbf{Random flows:} Each random flow includes six different transformations, and each transformation can repeat four times (example shown in Table \ref{fig:model_flows_24}), resulting in totally twenty-four transformations in each flow. These 100,000 random flows are generated by randomly permuting these twenty-four transformations.

\noindent
\textbf{Inputs and Labels:} The inputs of the neural network are the flows using the timed-model matrix representation with shape (6,24). The labels include the delay or area results collected by applying the random flow following by ABC technology mapping (command: \textit{map -v}). 

\begin{table}[!htb]
\scriptsize
\centering
\caption{Summary of Datasets. Data points are generated with 100,000 random flows using three different technology libraries using the first three designs. For the rest of the designs, the data points are collected with 20,000 random flows using the 7nm RVT and LVT FinFET technologies. The ground truth are the QoRs (delay or area) that are collected after technology mapping. \it{*RVT = Regular Voltage Transistor; *LVT = Low Voltage Transistor.}}
\label{tbl:datasets}
\begin{tabular}{|c|r|r|r|}
\hline
\textbf{Design} & \multicolumn{1}{c|}{\textbf{14nm}} & \multicolumn{1}{c|}{\textbf{7nm RVT}} & \multicolumn{1}{c|}{\textbf{7nm LVT}} \\ \hline
64-bit Montgomery & 99,997 & 99,997 & 99,997 \\ \hline
64-bit ALU & 100,000 & 100,000 & 100,000 \\ \hline
128-bit AES core & 99,737 & 99,737 & 99,737 \\ \hline
LU8PEng & - & 20,000 & 20,000 \\ \hline
Stereovison0 & - & 20,000 & 20,000 \\ \hline
Stereovison1 & - & 20,000 & 20,000 \\ \hline
SHA & - & 20,000 & 20,000 \\ \hline
raygentop & - & 20,000 & 20,000 \\ \hline
OR1200 & - & 20,000 & 20,000 \\ \hline
Boundtop & - & 20,000 & 20,000 \\ \hline
blob\_merge & - & 20,000 & 20,000 \\ \hline
bgm & - & 20,000 & 20,000 \\ \hline\hline
\multicolumn{2}{|c|}{\textbf{Inputs}} & \multicolumn{2}{c|}{\textbf{Ground truth}} \\ \hline
\multicolumn{1}{|c|}{Flow} & \multicolumn{1}{c|}{(6, 24)} & \multicolumn{1}{l|}{Delay/Area} & \multicolumn{1}{c|}{(1,1)} \\ \hline
\end{tabular}
\end{table}

\subsection{Training setups and pre-processing}\label{sec:training_setups}

\noindent
\textbf{Training setups:} The loss function is the \textit{mean squared error} (MSE) and is optimized with \textit{Adam} optimizer \cite{kingma2014adam} with \textit{learning rate}=0.001, $\beta_1$=0.9, $\beta_2$=0.999. The \textit{batch size} used in this work is 256 and models are trained for 1000 epochs.

\noindent
\textbf{Pre-processing (Data normalization):} The training data points are normalized before model training. 
Specifically, the label vectors are normalized by subtracting its \textit{mean} and dividing its \textit{range}. The \textit{mean} and \textit{range} are used to reconstruct the ground truth at testing. 

\section{Transferability cross Designs and Technologies}

\begin{figure}[!htb]
\centering
\includegraphics[width=0.44\textwidth]{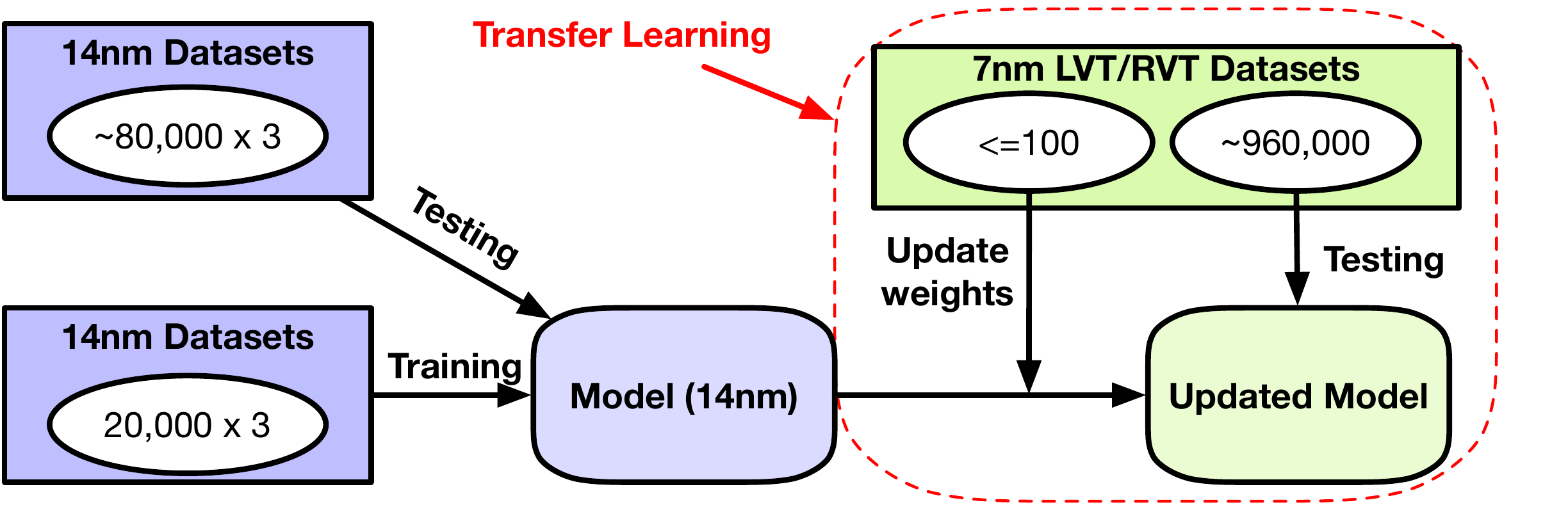}
\caption{Transfer-learning cross different technologies. Initial model is trained using 20\% of the 14nm datasets, i.e., 20,000 data points of each design. The pre-trained model is updated using $\leq$100 new data points, which are produced using unseen 7nm technologies and IC designs. The rest of the 7nm datasets are used for evaluating the transfer-learning approach, including $~$960,000 data points.}
\label{fig:transfer_learning_approach}
\end{figure}

We explore the transferability over different technologies and designs using the approach shown in Figure \ref{fig:transfer_learning_approach}. The main idea is to utilize the model pre-trained with 14nm data points and update the model with little data points to predict for unseen 7nm technologies and designs shown in Table \ref{tbl:datasets}. In this work, we restrict the number of data points for transfer learning to be $\leq$100. Specifically, the results of transfer learning using \{10, 25, 50, 100\} data points are included in the result section. The evaluations are made using the rest of the 7nm datasets.

\begin{figure*}[t]
\begin{minipage}{0.24\textwidth}
  \centering
\includegraphics[width=1\textwidth]{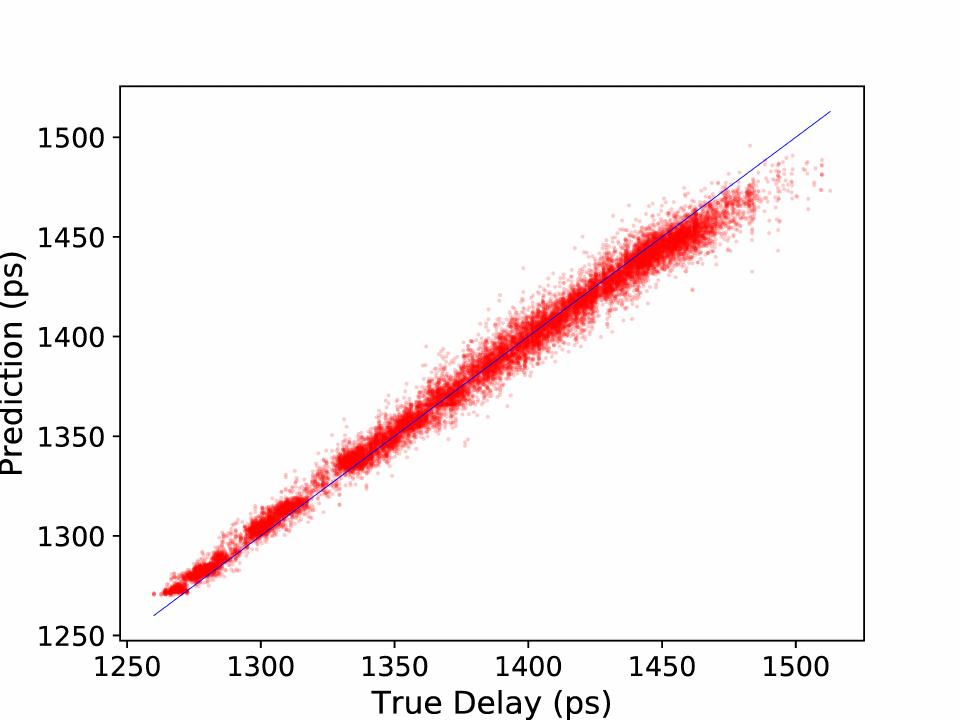}
\subcaption{Sub-set 1: avg error=0.370\%}
\end{minipage}%
\begin{minipage}{0.24\textwidth}
  \centering
\includegraphics[width=1\textwidth]{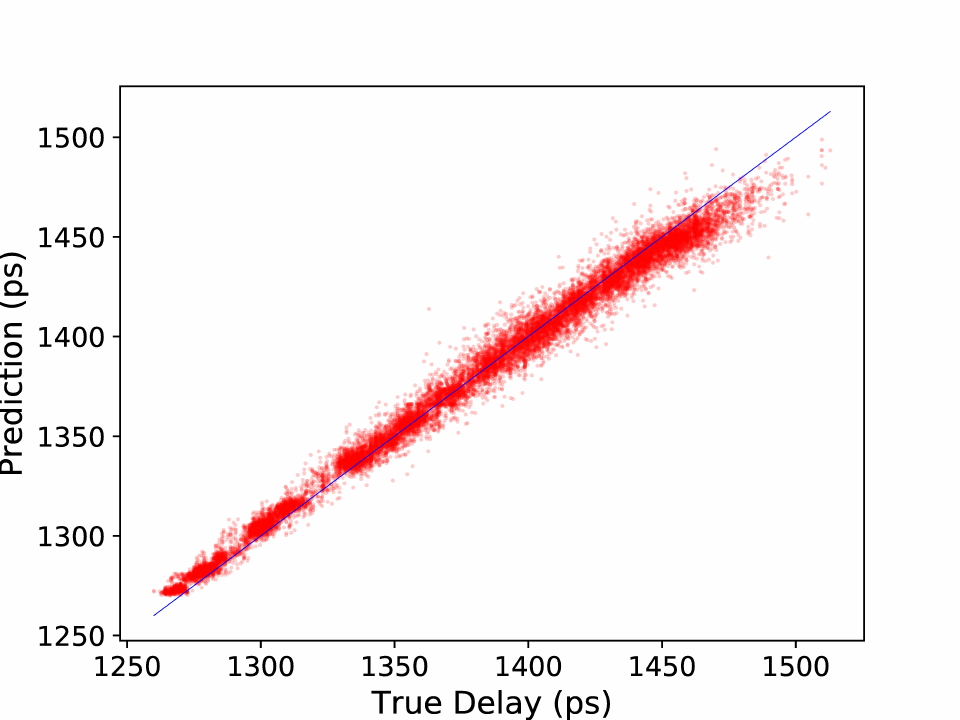}
\subcaption{Sub-set 2: avg error= 0.369\%}
\end{minipage}%
\begin{minipage}{0.24\textwidth}
  \centering
\includegraphics[width=1\textwidth]{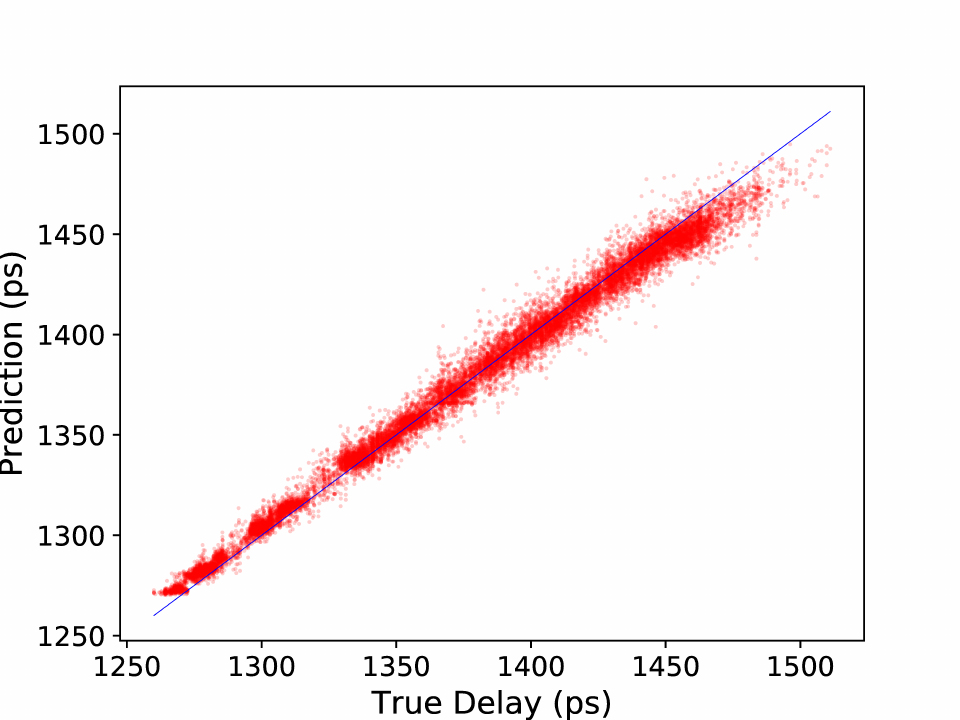}
\subcaption{Sub-set 3: avg error=0.367\%}
\end{minipage}%
\begin{minipage}{0.24\textwidth}
  \centering
\includegraphics[width=1\textwidth]{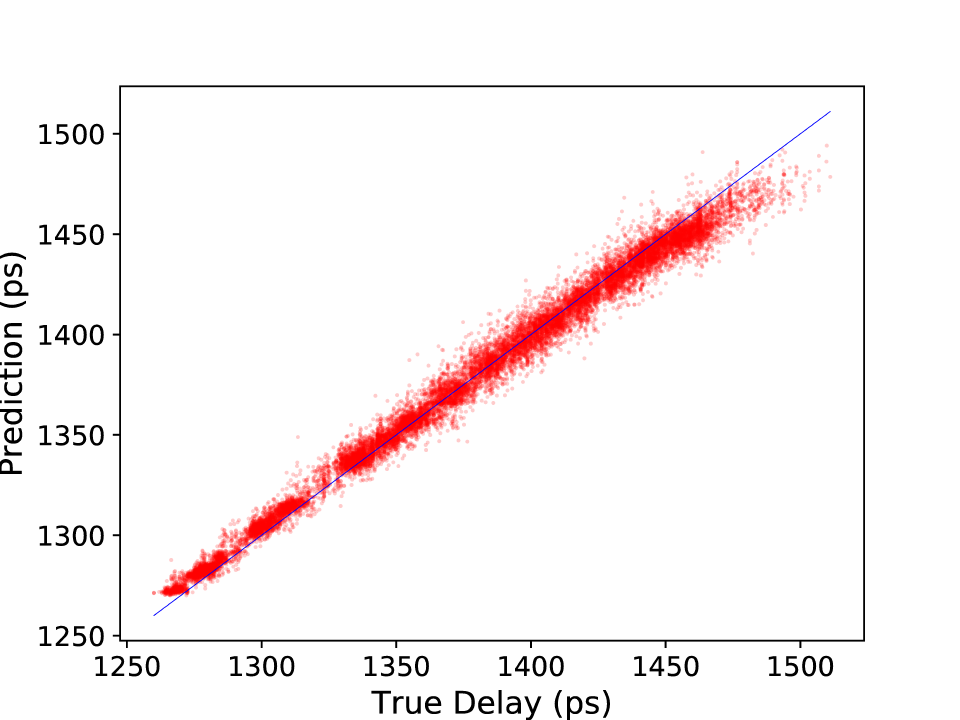}
\subcaption{Sub-set 4: avg error=0.371\%}
\end{minipage}
\caption{Visualization of delay prediction with the remaining $\sim$80,000 data points of 14nm Montgomery dataset as testing inputs. Each sub-set includes 20,000 test results (last sub-set includes 19,997). Overall prediction accuracy over 79,997 test points is 99.6\%.}
\label{fig:testing_14nm_GF}
\end{figure*}

\subsection{Transfer Learning Strategies}\label{sec:transfer_learning_approach}

Two transfer learning approaches are implemented. 

\noindent
\textbf{Updating Dense Layers:} This approach takes the pre-trained model and turns the LSTM layers to be non-trainable, i.e., layers 1-4 shown in Table \ref{tbl:table_rnn_model}. The main intuitions of this approach are that 1) the sequential behavior of the synthesis flows could be similar over different designs and technologies, and 2) the sequential features have been learned mostly in the LSTM layers. In this case, there are only about 5,000 parameters in the pre-trained model that need to be updated during transfer learning.

\noindent
\textbf{Updating All Layers:} However, if the sequential behaviors of the synthesis flows are different over different designs and technologies, the model could fail to converge without updating the LSTM layers. Hence, this approach updates the parameters of all layers in the pre-trained model.

\section{Result}

\begin{figure}[!htb]
\centering
\begin{minipage}{0.24\textwidth}
  \centering
\includegraphics[width=1\textwidth]{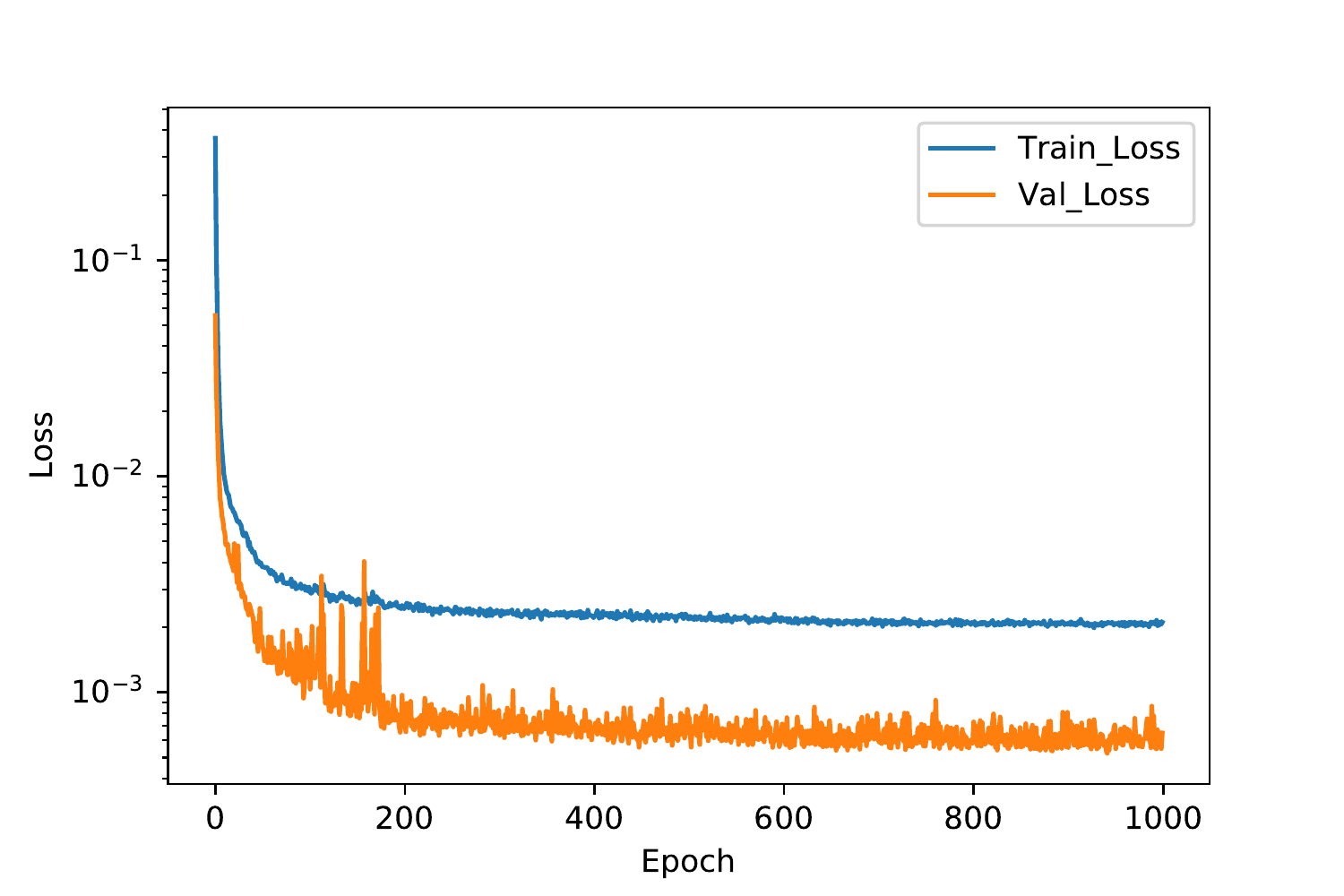}
\subcaption{Training loss vs. validation loss.}\label{fig:result_14nm_GF_1a}
\end{minipage}%
\hspace{0.15mm}
\begin{minipage}{0.225\textwidth}
  \centering
\includegraphics[width=1\textwidth]{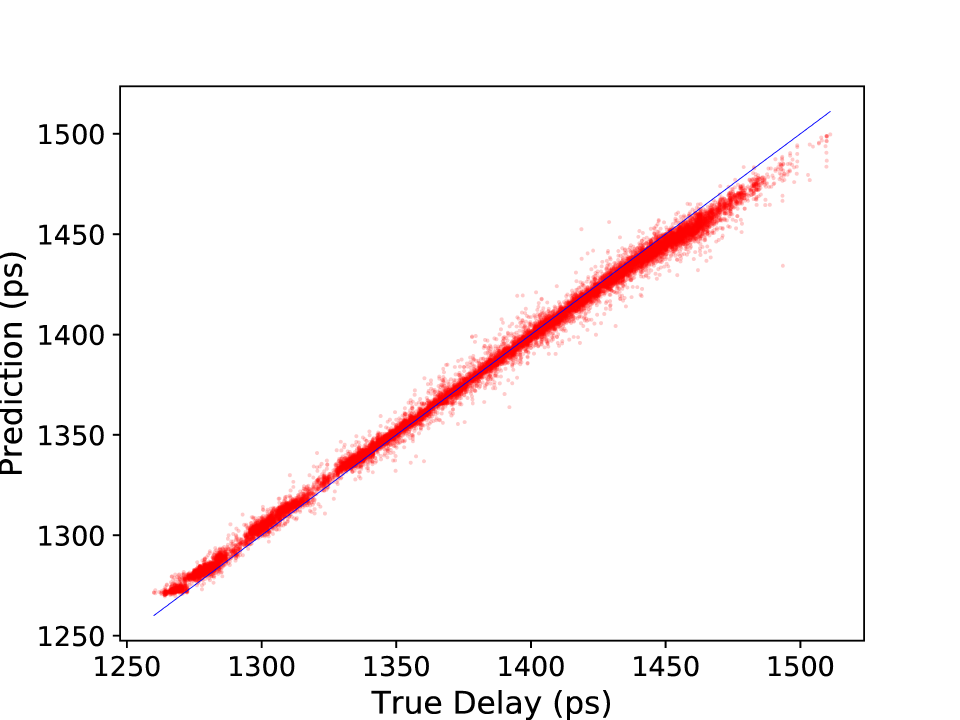}
\subcaption{Predictions using the training data.}\label{fig:fig:result_14nm_GF_1b}
\end{minipage}%
\caption{Training results with 20,000 14nm 64-bit Montgomery multiplier as inputs.}
\label{fig:result_14nm_GF}
\end{figure}

\begin{figure}[t]
\centering
\begin{minipage}{0.23\textwidth}
  \centering
\includegraphics[width=1\textwidth]{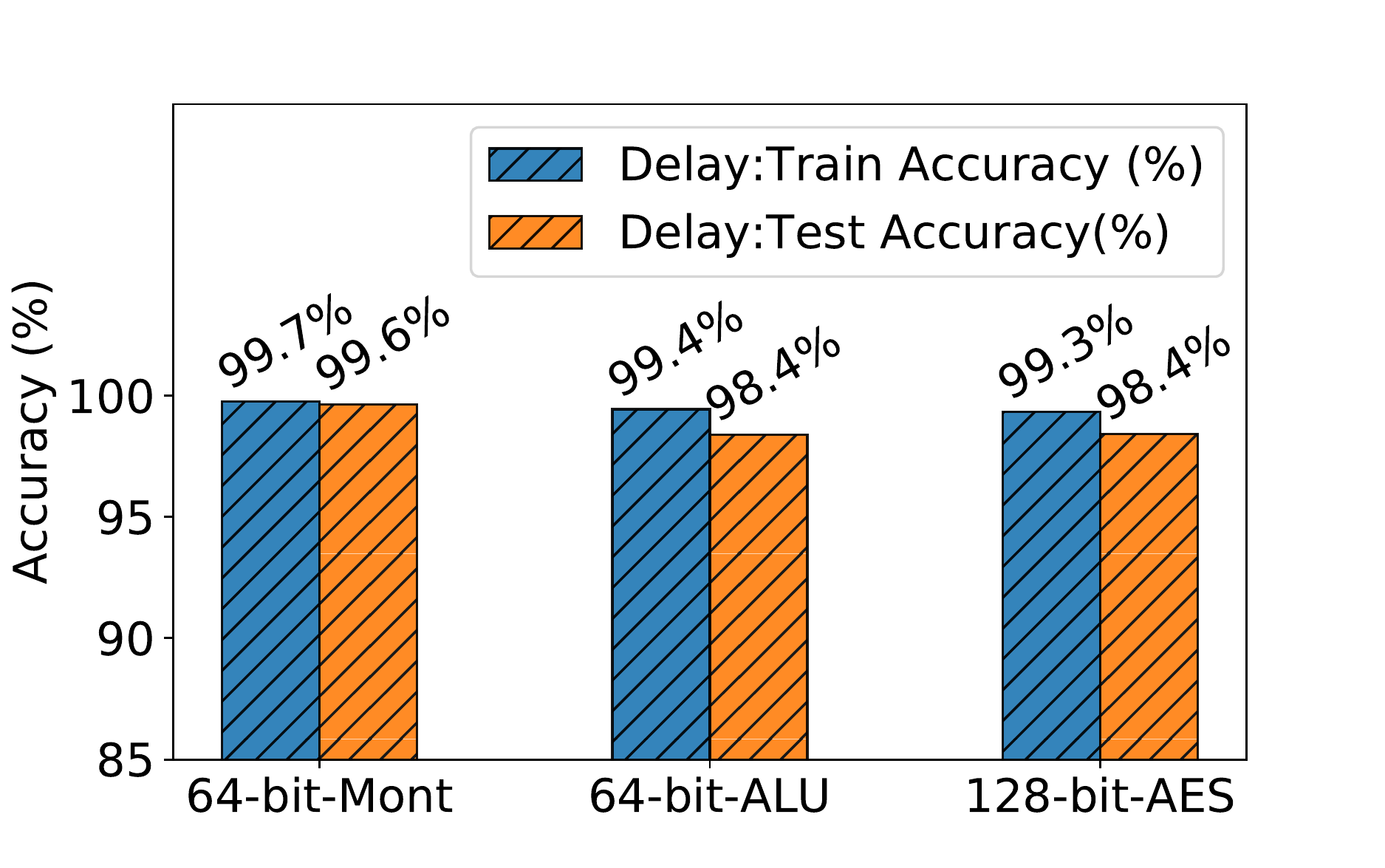}
\subcaption{Results of delay prediction.}\label{fig:14nm_evaluation_all_1a}
\end{minipage}%
\hspace{0.15mm}
\begin{minipage}{0.23\textwidth}
  \centering
\includegraphics[width=1\textwidth]{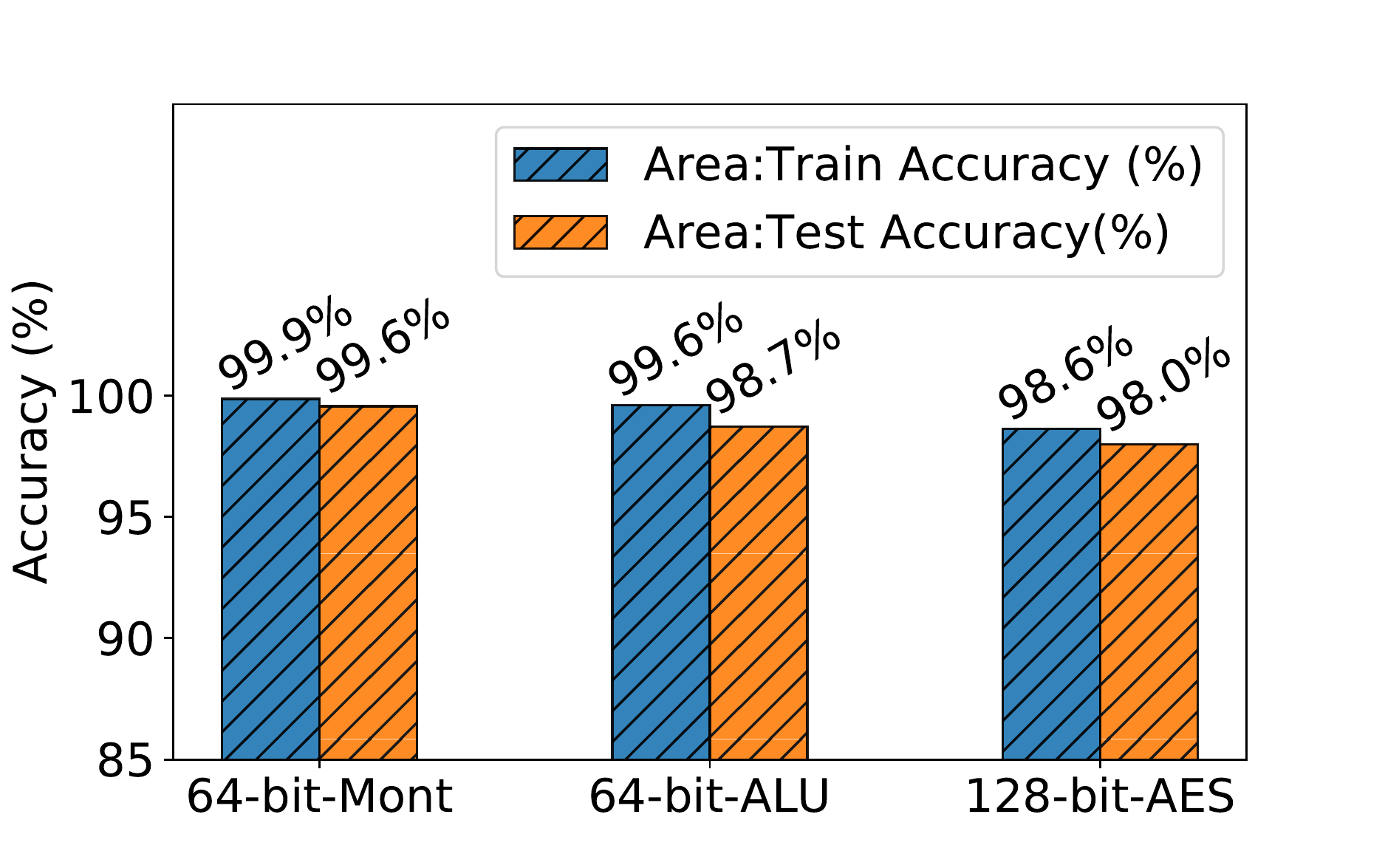}
\subcaption{Results of area prediction.}\label{fig:14nm_evaluation_all_1b}
\end{minipage}%
\caption{Evaluation of delay and area prediction using 14nm datasets.}
\label{fig:14nm_evaluation_all}
\end{figure}

First, the pre-trained model is evaluated with 14nm datasets by evaluating the delay and area prediction accuracy, with 20,000 data points for training and $\sim$80,000 for evaluation for each design (Table \ref{tbl:datasets}). Secondly, we evaluate our transfer learning approaches on the delay and area of the 7nm RVT/LVT datasets. The training and testing are conducted on a server with 28 Intel Xeon CPU E5-2690 v4 processors and 256 GB memory. The experiments are implemented in Python3 using deep learning framework Keras \cite{chollet2015keras} with Tensorflow \cite{abadi2016tensorflow} as backend.

\subsection{Evaluation within 14nm datasets}\label{sec:result_1}

The results in this section use the training setups provided in Section Approach, with epochs=1000, and 20\% training data for validation. The training results using the 14nm datasets (delay) of 64-bit Montgomery multiplier are shown in Figure \ref{fig:result_14nm_GF}. Specifically, Figure \ref{fig:result_14nm_GF_1a} shows the training loss and the validation loss ($y$-axis) with respect to the number of epochs ($x$-axis). Figure \ref{fig:fig:result_14nm_GF_1b} shows the prediction results with the flows of the training data points as inputs. The $x$-axis represents the true delay of the flow, and the $y$-axis represents the predicted delay, with unit \textit{ps}. The training data points fit perfectly after 1000 epochs, with average prediction error 0.255\%. 

The model is then tested using the remaining ~80,000 14nm data points. The testing results are shown in Figure \ref{fig:testing_14nm_GF}. The testing dataset is randomly split into four subsets, with each sub-set including $\sim$20,000 points. This is used to demonstrate that the prediction accuracy does not differ much while choosing different inputs. The average delay prediction error are 0.370\%, 0.369\%, 0.367\%, and 0.371\%, with an overall average error 0.369\%. Similarly, we evaluate our approach using the 14nm datasets for both delay and area, shown in Figure \ref{fig:14nm_evaluation_all}. The y-axis is the training and testing accuracy for delay and area of three designs. The delay testing accuracy are 99.6\%, 98.4\%, and 98.4\% (Figure \ref{fig:14nm_evaluation_all_1a}), and the area testing accuracy are 99.6\%, 98.7\%, 98.0\% (Figure \ref{fig:14nm_evaluation_all_1b}). In summary, $\sim$80,000 test data points, the prediction accuracy for delay and area for all three designs are $\geq$ 98.0\%.

\subsection{Transfer Learning: cross designs and technologies}\label{sec:result_2}


This section presents the evaluation results of transfer learning using the approaches shown in Figure \ref{fig:transfer_learning_approach}. The testing datasets are the 7nm datasets shown in Table \ref{tbl:datasets}. We first show the complete delay prediction results using 64-bit GF multiplier design. The initial model is pre-trained with 20,000 14nm data points. Then, we update the weights of all layers of this model with 100 7nm data points from the same design. The rest data points for each technology are used for testing. The prediction accuracy is $\geq$99.5\% for both 7nm technologies. Note that these show the transferability cross the technologies only.

\begin{figure}[!htb]
\centering
\begin{minipage}{0.242\textwidth}
  \centering
\includegraphics[width=1\textwidth]{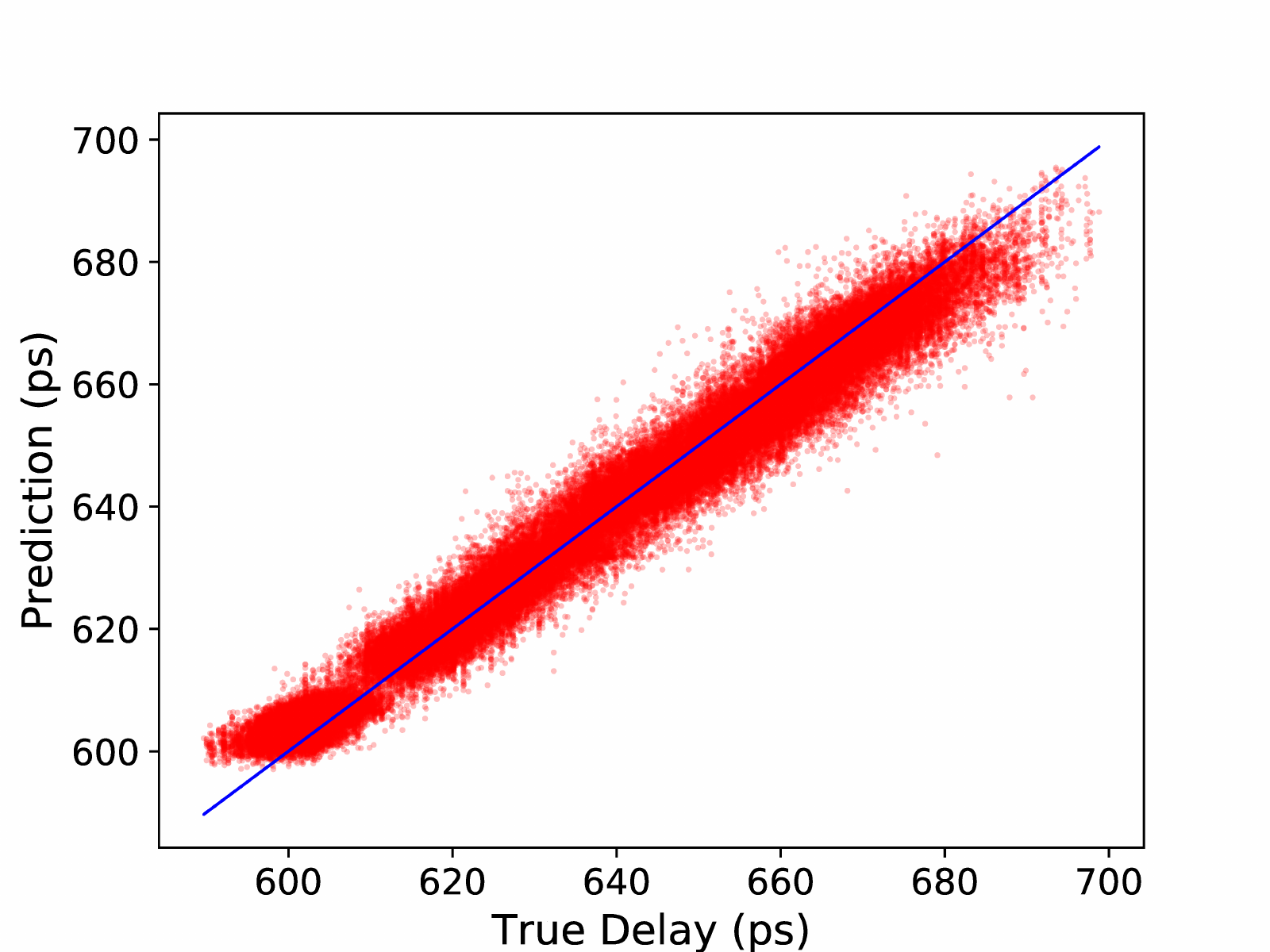}
\subcaption{100k 7nm RVT predictions. \\Average accuracy = 99.5\%}
\end{minipage}%
\begin{minipage}{0.242\textwidth}
  \centering
\includegraphics[width=1\textwidth]{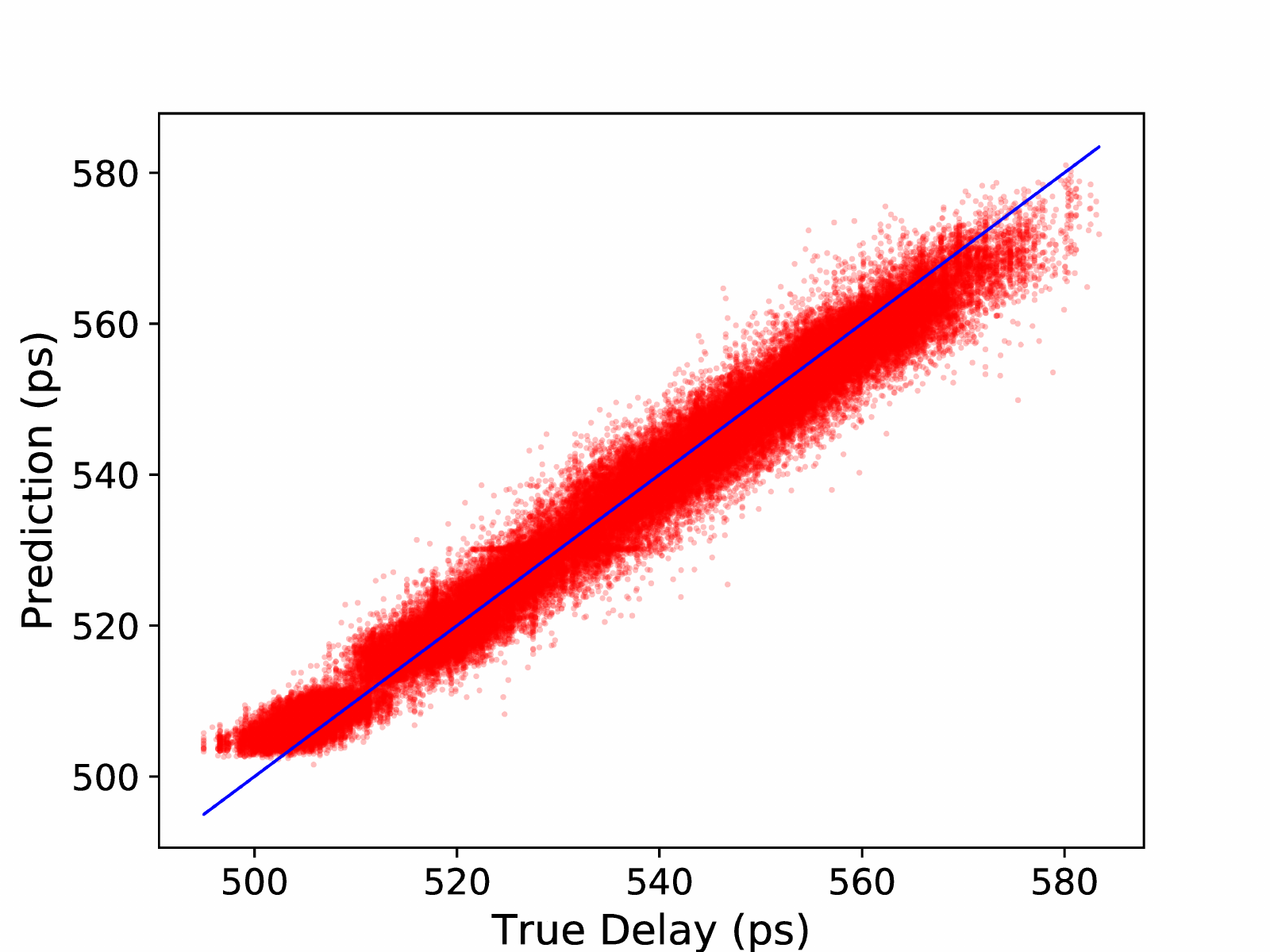}
\subcaption{100k 7nm LVT predictions. \\Average accuracy = 99.6\%}
\end{minipage}%
\caption{Visualization of 7nm delay predictions using transfer learning with prediction accuracy $\geq$99.5\%. Initial model is trained with 64-bit GF multiplier 14nm datasets, and is updated with 100 7nm data points.}
\vspace{-2mm}
\label{fig:transfer_GF}
\end{figure}

\begin{figure}[!htb]
  \centering
  \begin{minipage}{0.48\textwidth}
    \centering
\includegraphics[width=1\textwidth]{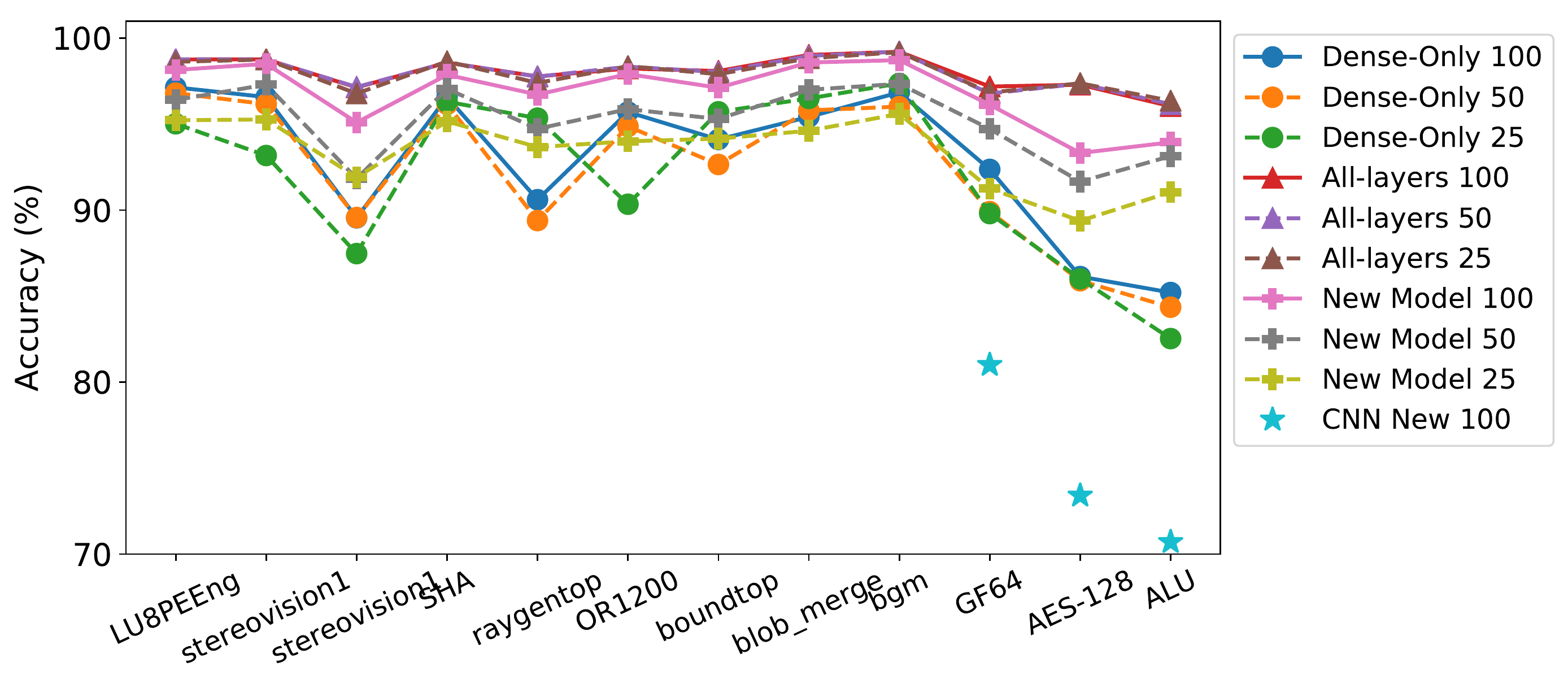}
\subcaption{Delay.}\label{fig:transfer_delay}
\end{minipage}%
\\
  \begin{minipage}{0.48\textwidth}
    \centering
\includegraphics[width=1\textwidth]{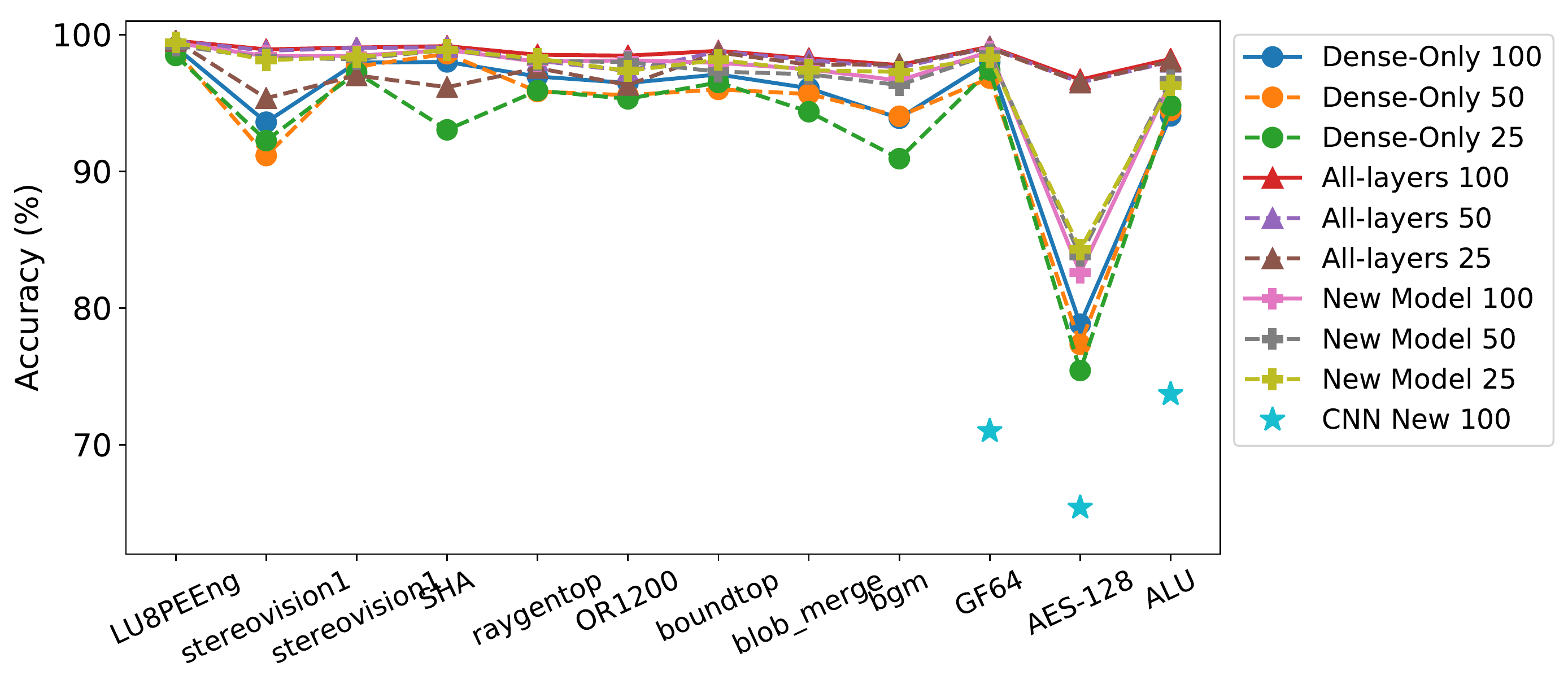}
\subcaption{Area.}\label{fig:transfer_area}
\end{minipage}%
\caption{Evaluation of two transfer learning approaches using 25/50/100 data points. \textit{CNN New 100} results are generated using technique in \cite{cunxi2018CNN} with 100 data points for transfer learning.}
\vspace{-2mm}
\label{fig:transfer_learning_result}
\end{figure}

To explore the transferability cross both designs and technologies, we apply transfer learning to all 12 designs with the initial model pre-trained with 14nm GF delay dataset. Note that industrial studies indicate that machine learning based electronic design systems require a minimum of 95\% accuracy for performance estimation \cite{mentor18dac}. More importantly, these systems are required to be stable (i.e., have similar accuracy) for different types of designs. The results of two approaches, updating dense layers only and updating all layers, are included in Figure \ref{fig:transfer_learning_result}. To demonstrate the advantages of transfer learning, the results of training a new model from scratch using the same amount of data points, \textit{without} pre-training on 14nm data first, are included as a baseline. It shows that transfer learning by updating all layers provides the best results over all designs, for delay and area estimations. Our approach obtains $\geq$96.3\% accuracy for delay and area over all designs, using 100 7nm data points.
In comparison to transfer learning approaches, training a new model without transfer learning yields much worse accuracy due to insufficient training data. This indicates that transfer learning is helpful when the size of available training data is limited.

We also compare our LSTM network with the CNN based approach. We modify the model released in \cite{cunxi2018CNN} by 1) changing the the output layer 
from multi-channel \textit{softmax} output to single-channel \textit{linear} output
and 2) adding BN following the Convolutional layers. For fair comparison, we only compare the delay/area estimation accurate for GF, AES, and ALU designs that were used in that work. It shows that with 100 training data points, CNN regressor performs much worse than the LSTM regressor (Figure \ref{fig:transfer_learning_result}). 
The transfer-learning results using CNN approach are not included since they perform worse than training new CNN model.

\begin{figure}[!htb]
\centering
\includegraphics[width=0.44\textwidth]{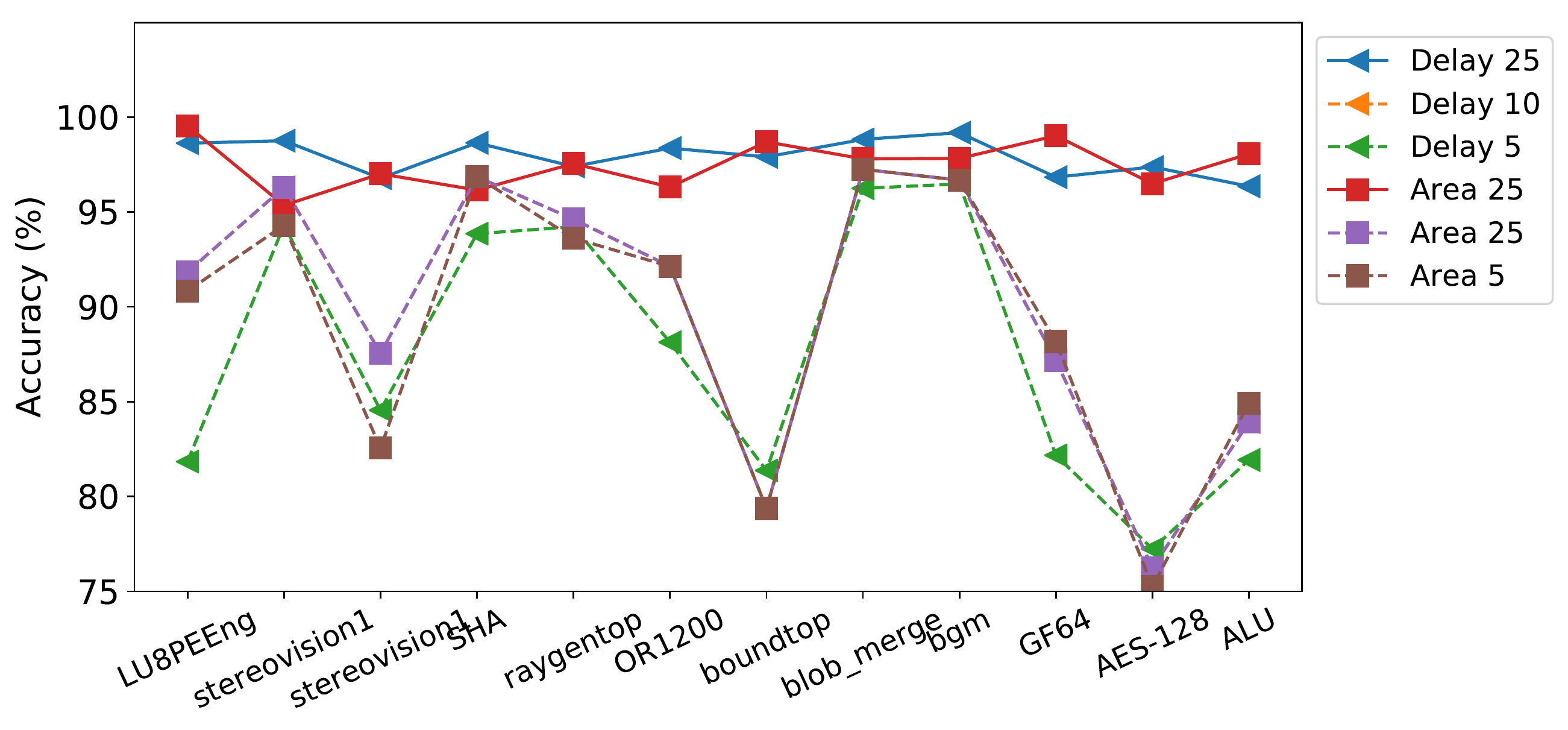}
\caption{Evaluation of transfer learning (updating all layers) for delay and area estimation using 5/10/25 data points.}
\vspace{-2mm}
\label{fig:minimum_data_points}
\end{figure}

Finally, we try to find the minimum number of data points for transfer learning to achieve reasonable accuracy. Specifically, we choose the approach of updating all layers, and set the number of training data points to 5/10/25. The results are shown in Figure \ref{fig:minimum_data_points}. For both delay and area, the estimation accuracy significantly decreases with $\leq$10 data points. This suggests that at least 25 data points are needed for the proposed transfer learning approach to achieve stable estimation accuracy cross different designs and technologies.

\section{Conclusion}
This paper presents an RNN regression based approach that precisely estimates the delay and area of synthesis flows. The proposed RNN regressor is constructed using LSTM network with batch normalization and dense layers. To enable accurate predictions for future technologies and different designs, we propose a transfer-learning approach that utilizes the pre-trained model and requires much less training data. The demonstrations are made with logic synthesis tool ABC using 14nm and 7nm FinFET technologies, and models are tested over 1.2 million data points. The results show the prediction accuracy of delay and area is $\geq$98.0\% for single technology, and the prediction accuracy after transfer-learning cross designs and technologies is $\geq$96.3\% with only 100 new data points. This demonstrates that the proposed transfer learning approach can effectively learn to estimate QoR for unseen technologies and designs. Future work will focus on performance estimations at physical layout level (e.g., silicon routing congestion), and 5nm technologies.

\small
\bibliographystyle{IEEEtran}
\bibliography{synthesis}

\end{document}